\pdfoutput=1
\documentclass{article} 
\usepackage{iclr2021_conference,times}

\usepackage{hyperref}
\usepackage{url}

\usepackage{algorithm}
\usepackage{xcolor}
\usepackage{color}
\usepackage{algorithmic}
\usepackage{graphicx}
\usepackage{float}
\usepackage{subfigure}
\usepackage{colortbl}
\usepackage{enumitem}
\definecolor{Gold}{RGB}{191, 144, 0}

\setlist[itemize]{leftmargin=*}
\setlist[enumerate]{leftmargin=*}
\newcommand{\method}{\textsc{Self-Familiarity}}
\newcommand{\data}{Concept-7}
\newcommand{\inference}{Concept Guessing}
\usepackage{amsmath}
\usepackage{multirow}
\title{Zero-Resource Hallucination Prevention for Large Language Models}

\author{Junyu Luo \\
Pennsylvania State University\\
\And
Cao Xiao \\
GE HealthCare \\
\AND
Fenglong Ma \\
Pennsylvania State University\\
}

\iclrfinalcopy 
\begin{document}

\maketitle

\begin{abstract}
The prevalent use of large language models (LLMs) in various domains has drawn attention to the issue of "hallucination," which refers to instances where LLMs generate factually inaccurate or ungrounded information. Existing techniques for hallucination detection in language assistants rely on intricate fuzzy, specific free-language-based chain of thought (CoT) techniques or parameter-based methods that suffer from interpretability issues. Additionally, the methods that identify hallucinations post-generation could not prevent their occurrence and suffer from inconsistent performance due to the influence of the instruction format and model style. In this paper, we introduce a novel pre-detection self-evaluation technique, referred to as {\method}, which focuses on evaluating the model's familiarity with the concepts present in the input instruction and withholding the generation of response in case of unfamiliar concepts. This approach emulates the human ability to refrain from responding to unfamiliar topics, thus reducing hallucinations. We validate {\method} across four different large language models, demonstrating consistently superior performance compared to existing techniques. Our findings propose a significant shift towards preemptive strategies for hallucination mitigation in LLM assistants, promising improvements in reliability, applicability, and interpretability\footnote{code and data can be found in \url{https://github.com/soap117/Self-evaluation}}.
\end{abstract}



\section{Introduction}

The widespread adoption of Large Language Models (LLMs) has generated significant interest in their application across diverse use cases, e.g., healthcare and medicine. However, a major challenge hindering their full potential is the issue of \textbf{hallucination}, where the models produce inaccurate or fabricated information, leading to a substantial gap in their reliability and trustworthiness.
Figure~\ref{fig:example} shows an example to demonstrate the problem of hallucination when a user queries ChatGPT about the drug ``\textit{Skytrofa}'' that aims to treat short stature due to inadequate secretion of endogenous growth hormone.\footnote{\url{https://www.accessdata.fda.gov/drugsatfda_docs/label/2021/761177lbl.pdf}} We can observe that ChatGPT generates inaccurate responses on essential factors of this drug.

Recently, several approaches have been proposed to detect (and even correct) the hallucination responses generated by LLMs in open conversations, which can be roughly classified into two categories: 
The first category of methods hinges on a third-party knowledge base, combined with the \emph{Chain-of-Thought} (CoT) technique, to identify and rectify potential hallucination responses~\cite{gou2023critic,zhang2023mitigating,min2023factscore,zhang2023interpretable, mundler2023self}. 
The other research direction focuses on designing \emph{parameter-based methods}~\cite{manakul2023selfcheckgpt,min2023factscore}, which mainly utilize specific metrics such as perplexity to evaluate the correctness of responses.

\begin{figure}[t]
\centering
\includegraphics[width=0.6\textwidth]{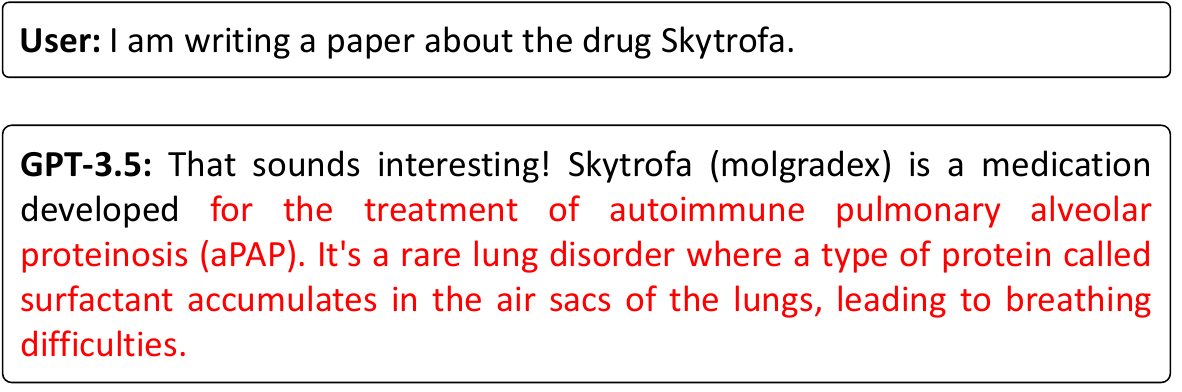}
\caption{A hallucination example. \textcolor{red}{Red} color indicates the incorrect information.}
\label{fig:example}
\end{figure}



All the aforementioned methods primarily focus on the \textbf{post-detection} of hallucination responses. These approaches can merely ascertain if a response is a hallucination, lacking the capacity to inhibit the production of such responses in the future, thereby diminishing reliability. Moreover, the performance of existing methods is deeply influenced by instruction and model styles, leading to challenges in maintaining robustness across open conversation scenarios. This complexity hinders the ability to establish a clear threshold for distinguishing hallucination responses. For instance, binary queries often prompt brief response sequences, leading to metrics that diverge significantly from those derived from more extensive response sequences. Therefore, a proactive, preventative strategy for hallucination responses is essential for practical and efficient applications of artificial intelligence (AI) language assistants.

Designing such an effective prevention method faces several challenges. 
Firstly, the proposed approach must navigate a \textbf{zero-resource} environment, precluding any reliance on external knowledge gleaned from search engines. Neglecting this imperative compromises the method's universality and applicability, rendering it unsuitable for situations with budgetary constraints or contexts lacking external access. Consequently, a profound comprehension of the language model's intrinsic knowledge becomes essential.
Furthermore, the task of ensuring robustness is of paramount significance. The envisaged system must exhibit resilience against diverse instruction types, contextual variations, and model styles. Given the open-ended and dynamic nature of human language, achieving consistent performance and unwavering resilience across a wide array of scenarios presents an undeniably formidable challenge.




To tackle these challenges simultaneously, we propose a novel zero-resource, pre-detection method named {\method}, illustrated in Figure~\ref{fig:procedure}. This approach mimics human self-assessment capabilities by refraining from discussing unfamiliar concepts, thereby reducing the risk of creating hallucinated information. This method sets it apart from conventional post-detection techniques.
Initially, our method extracts and processes concept entities from the instruction during the \textbf{Concept Extraction} stage. Following this, the \textbf{\inference} stage individually examines the extracted concepts through prompt engineering to obtain each concept's familiarity score. Lastly, during the \textbf{Aggregation} stage, the familiarity scores from different concepts are combined to generate the final instruction-level familiarity score.

Compared to existing methods, our algorithm presents the following advantages. Primarily, {\method} integrates the strengths of both CoT techniques and parameter-based methods. Like the CoT methods, our algorithm can offer constructive responses by identifying concepts unfamiliar to the model. Yet, our algorithm solely employs prompt engineering, eliminating the need for the model to possess strong inference abilities and avoiding their shortcomings while combining their advantages. Additionally, our algorithm remains unaffected by instruction style and type and is proactive and preventative, thereby increasing its reliability and robustness. Finally, it does not require any outside knowledge.

We assessed our method across four large language models using a newly proposed pre-detection hallucinatory instruction classification dataset, {\data}. Experimental results show that the proposed {\method}\footnote{The data and code can be found in \textit{Supplementary Material}.} consistently outperforms other methods across all models, demonstrating its huge application value.



\section{Related Work}\label{sec:rel_dataset}
To the  best of our knowledge, no existing work has been devoted to preventing hallucinated responses in open conversations by analyzing the instruction itself under the zero-resource setting. Consequently, the contexts we address are distinct from those of previous studies.

\subsection{Hallucination Detection and Correction Methods}
Previous studies in hallucination detection and correction mainly concentrated on conditional text generation for specific tasks such as abstract summarization~\cite{maynez2020faithfulness, wang2020asking, cao2021hallucinated}, image captioning~\cite{rohrbach2018object,biten2022let}, dialogue generation~\cite{shuster2021retrieval}, machine translation~\cite{zhou2020detecting, wang2020exposure}, and table-to-text generation~\cite{wang2020towards, wang2021sketch}. Since these works are highly task-specific, they fail to tackle hallucination issues in open conversations.

For an open conversation setting, the methodologies are typically categorized into two groups based on the employed strategies. The first group utilizes the Chain of Thoughts (CoT) or prompt programming to evaluate and amend responses~\cite{lee2022factuality,gou2023critic,zhang2023mitigating,min2023factscore, peng2023check,huang2023zero,xie2023adaptive,yue2023automatic}. A noteworthy example is CRITIC~\cite{gou2023critic}, wherein a CoT process is deployed with supplementary inputs from an external search engine to enhance response quality. Certain works do not necessitate external knowledge~\cite{zhang2023interpretable, mundler2023self}, often directly asking the model to assess the output's faithfulness. Nonetheless, such methods can be limited as they are engineered for specific responses, and highly depend on the inner inference ability of the model. Another challenge lies in the fact that the algorithm output is usually free text, which can make the actual classification threshold ambiguous.

The second category of methods~\cite{manakul2023selfcheckgpt, min2023factscore} emphasizes using language model parameters, such as token probability sequence, to determine the hallucination level. These methods generally exhibit superior generalization capability and can provide precise output scores. Self-check GPT~\cite{manakul2023selfcheckgpt} pioneers the use of parameter-based methods for open-ended text generation. In Self-check GPT, the perplexity, sampling, and unconditional probability are used together to estimate the hallucination level. However, this work only assesses biography-related issues, and compared to CoT techniques, the model's interpretability is significantly reduced.

\begin{figure*}[t]
\centering
\includegraphics[width=1.0\textwidth]{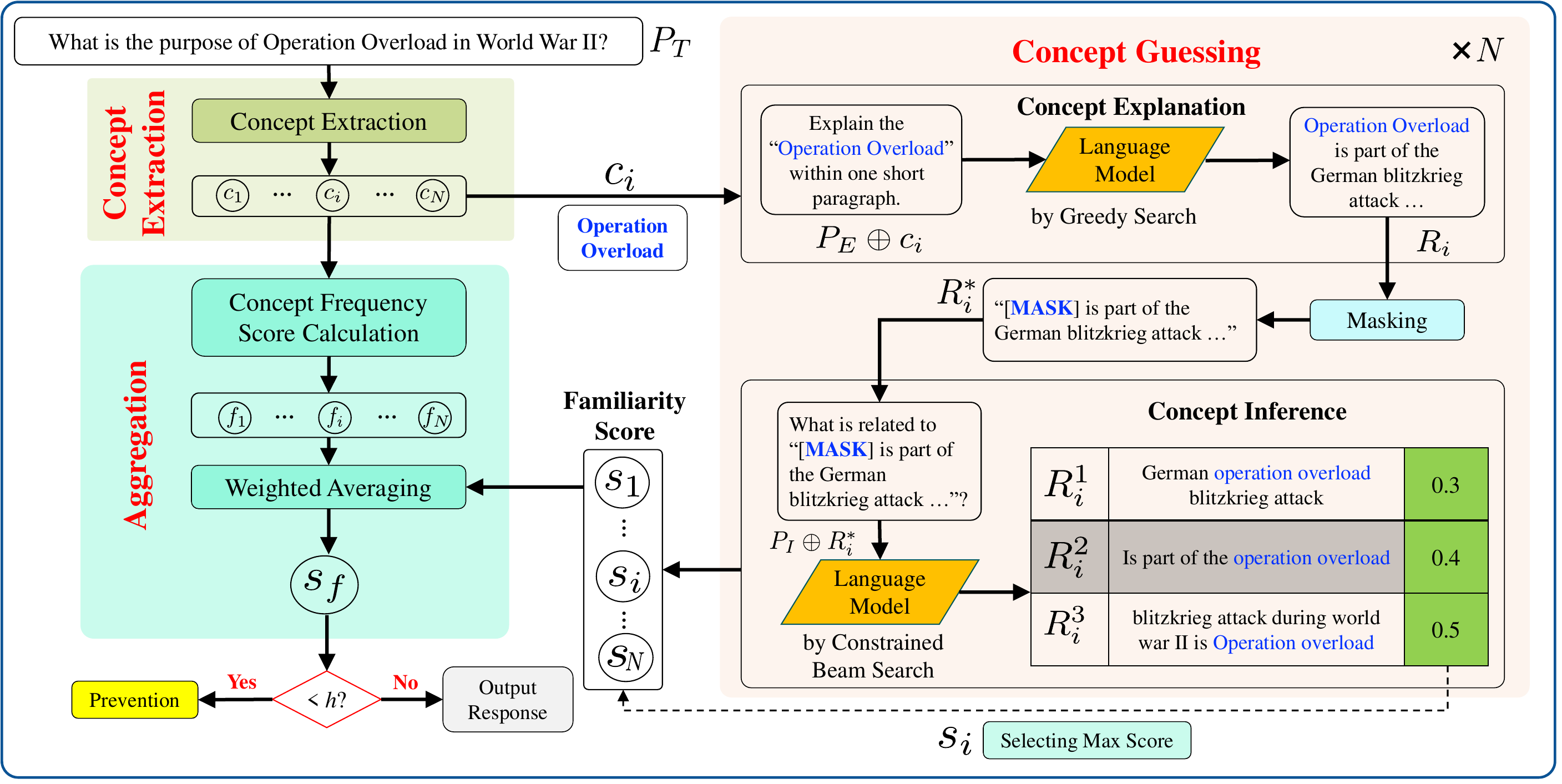}
\caption{Example procedure of the {\method}. 
}
\label{fig:procedure}
\end{figure*}

\subsection{Hallucination Detection Datasets}
Current datasets for hallucination detection in open conversations majorly focus on post-detection scenarios. In these datasets~\cite {lin2021truthfulqa,liu2022token, muhlgay2023generating, li2023halueval, manakul2023selfcheckgpt, min2023factscore, mundler2023self, zhang2023interpretable}, the task involves selecting the correct response or ascertaining whether responses are incorrect. However, these datasets are subject to certain constraints. Primarily, the datasets often originate from a single task like biography writing, without accounting for variations across different language models. Each model can have distinct background knowledge, enabling one model to identify a hallucinated response that another model might overlook. Furthermore, even if a model can accurately classify a specific hallucinated response, it does not guarantee that the model will refrain from generating a different hallucinated response in the future. Therefore, it is important to create a new dataset used for validating the pre-detection setting.

\section{Methodology}

The aim of our algorithm is to evaluate if the target instruction $P_T$ is a potential hallucinatory instruction by checking the familiarity of the language model with the concepts that exist in the $P_T$ under the zero-resource setting.
Our method, as depicted in Figure~\ref{fig:procedure}, comprises three major steps: (1) Concept Extraction, (2) {\inference}, and (3) Aggregation.
The details of each step will be elucidated in the subsequent sections.

\subsection{Concept Extraction} 
To assess familiarity, it is necessary to first extract the concept entities from the free-text instruction, otherwise, the score will be greatly influenced by the ``\textit{noise},'' i.e., the stylistic and formatting elements of the instruction that do not contribute to its understanding. For example, the transformation of the question \textit{``Can sound travel in a vacuum?''} into the statement \textit{``Sound can travel in a vacuum. Please judge the statement.''} doesn't alter the knowledge requisite, but it can substantially modify the style of the subsequent response. In addition, if the instructions contain multiple concepts, it will greatly increase the difficulty of the following prompt engineering.
By isolating and individually evaluating these concepts, we can minimize such stylistic influence, thereby enhancing the robustness of subsequent procedures. We achieve this extraction through the utilization of a Named Entity Recognition (NER) model on the given instruction:
\begin{align}
    [c_1,\cdots, c_{N}] = \mbox{NER}(P_T).
\end{align}
We consider these extracted entities $[c_1,\cdots, c_{N}]$ to be the key concepts of the instruction. $N$ stands for the number of extracted concepts. However, NER models frequently produce extraneous noise and fail to completely capture some concepts. Consequently, we introduce subsequent processing steps to refine these extracted concepts, as outlined in the following sections.

\emph{Concept Grouping}. 
The extracted concepts frequently exhibit a degree of incompleteness. For instance, the term ``2023 United States debt-ceiling crisis'' could be inadvertently divided into [``2023'', ``United States'', ``debt-ceiling crisis'']. To address this issue, we propose to sequentially fuse the adjacent concepts. We sort the concepts based on their position in $P_T$ and attempt to fuse one concept with the adjacent concept, if the newly combined concept is found within the original $P_T$, we integrate the original pair to create an extended, unified concept. 

\emph{Concept Filtering}.
After merging the concepts, the subsequent step entails the exclusion of simple concepts that should not be examined to improve efficiency. 
These include common concepts such as ``year'' and ``age'', which are generally well comprehended by the models.
To address this, we identify the top frequently used words in Wiktionary\footnote{\url{https://en.wiktionary.org/wiki/Wiktionary:Main_Page}}, which we designate as ``common concepts.'' Any concepts encompassed within these common words are subsequently eliminated.

\subsection{{\inference}}
Our next task is to examine the familiarity of the Language Model (LM) with the extracted concepts in a zero-resource setting. Undertaking this task in a zero-resource setting magnifies its complexity, as it precludes reliance on external concept knowledge. As such, comparing the model's understanding with an established gold definition to derive results becomes an impracticable approach. Meanwhile, a direct query to the model regarding its familiarity with specific methods, such as CoT, presents one possible solution. However, this approach demands a robust inference capability on the part of the model and frequently results in ambiguous responses. Consequently, there is a need for a more universally applicable and precise method.

In response to this need, we introduce a novel self-evaluation technique, denoted as {\inference}. Our approach begins by prompting the model to generate an explanation for a given concept. Subsequently, through prompt engineering, we ask the model to recreate the original concept based on this explanation. Should the model generate the initial concept successfully, the probability score of the response sequence can be interpreted as the connection strength between the concept and the explanation, serving as a familiarity score.
This entire process can be likened to a specialized Charades or Pictionary game. If a language model can proficiently derive the original concept from its generated explanation, this not only suggests the sufficiency of the explanation but also reflects the model's adeptness with the concept. Importantly, our method does not necessitate the acquisition of a gold definition of the concept.
We outline the following steps to transform this conceptual approach into a standardized metric:

\subsubsection{Concept Explanation}
First, we use a standard explanation prompt $P_E$ in conjunction with the target concept $c_i$ to query the tested LM. This inquiry prompts the LM to generate an explanation for each concept through greedy search (GreedySearch), which selects the next word with the highest probability to generate the response. This process continues until it encounters the end of the sentence token or reaches the maximum defined length, denoted as $l_F$:
\begin{align}
    R_i = \mbox{GreedySearch}(\mbox{LM}(P_{E}\oplus c_i)),
\end{align}
where $\oplus$ denotes inserting $c_i$ into the pre-defined position of $P_E$.
In many scenarios, the original concept may be directly incorporated into the generated explanation, as illustrated in Figure~\ref{fig:procedure}. Consequently, the model could simply ``copy'' the original concept to ``cheat''. To prevent this, we mask the words of the $c_i$ within the $R_i$:
\begin{align}
    R^*_i = \mbox{Mask}(R_i).
\end{align}

\subsubsection{Concept Inference}
Given the masked explanation $R^*_i$ and the concept inference prompt represented by $P_{I}$, we can ask the model to generate the original concept $c_i$. However, the response from the model is produced as open-ended free text, which poses a challenge when attempting to transform it into a standard score.
Consider an instance where the model might correctly generate the original concept, but in a different format such as \textit{``Coca-Cola's biggest competitor''} rather than \textit{``Pepsi''}. This discrepancy complicates the determination of whether the original concept has been successfully regenerated.
To resolve this, we apply the constrained beam search (ConsBeamSearch)~\cite{anderson2017guided} approach, instructing the model to seek responses incorporating the original concept through beam search and providing the probability score of the responses in the meanwhile:
\begin{align}
\begin{split}
    &[< R^1_i, s^1_i >,\cdots , <R^{T_B}_i, s^{T_B}_i>]\\ =& \mbox{ConsBeamSearch}(\mbox{LM}(P_{I}\oplus R^*_i), c_i),
\end{split}
\end{align}
where each $< R^j_i, s^j_i >$ corresponds to a response $R^j_i$ including the concept entity $c_i$\footnote{We consider the lowercase, uppercase, and capitalized forms in searching.}, with $s^j_i$ representing the corresponding response probability score. $T_B$ is the beam search size of the ConsBeamSearch algorithm. 
We set the stopping criteria to hit the end of the sentence token or reach the maximum length $l_B$.
Beam search will return multiple results, however, the understanding of the model is only related to the highest one.
Therefore, we choose the highest response score, $s_i$, from $[s^1_i,\cdots, s^{T_B}_i]$ as the familiarity score of the concept $c_i$:
\begin{equation}
    s_i = \mbox{Max}(s^1_i,\cdots, s^{T_B}_i).
\end{equation}
\subsection{Aggregation}
In many situations, the number of final extracted concepts can be greater than one. As a result, we need to rank the importance of each concept and merge the familiarity scores of the concepts based on their importance to generate a final, instruction-level outcome.

\subsubsection{Concept Frequency Score}
In order to evaluate the importance of a concept, we propose a method for calculating a score based on the frequency rank of words contained within that concept. Our expectation is that the concept with more infrequent words will correspond to a lower score $f_i$.
To do so, we retrieve the frequency rank $p^j_i$ of the $j$-th word of concept $c_i$ from the Wiktionary. We set the index to the length of the dictionary if the word is outside the dictionary or is capitalized.
Given that word distribution tends to follow a long-tail distribution, we employ the exponential function to transform the frequency rank back to a frequency score and multiply them together to obtain the concept level frequency score:
\begin{equation}
    f_i = \prod_{j=1}^{M_i}e^{\frac{-p_i^j}{H}}.
\end{equation}
Here, $M_i$ is the number of words in concept $c_i$. The term $H$ is introduced as a normalization factor to guarantee that the resulting score resides within a reasonable range.

\subsubsection{Weighted Aggregation}
Next, we average the familiarity scores based on their frequency scores through a weighted average. This approach offers robustness in multi-entity situations when compared to simply selecting a single score as the final value. To make the important parts contribute more than the less important tail parts, we establish a geometrically decreasing weighting scheme with a ratio of $\frac{1}{r}$:
\begin{equation}
    s_{f} = \frac{\sum_{i=1}^{N}{(r^{\theta(f_i)})}^{-1}s_i}{\sum_{i=1}^{N}{(r^{\theta(f_i)})}^{-1}}.
\end{equation}
Here, $\theta(f_i)$ denotes the rank position of the $f_i$ within $[f_1, \cdots, f_N]$ when sorted by magnitude of $f_i$.
We utilize the derived $s_{f}$ to represent the hallucination level of the instruction and terminate the response process if the score falls below the predetermined threshold $h$.

\section{Experiments}
In this section, we introduce the experimental settings and results. \emph{Due to space limitations, we have included the \textbf{implementation details}, and \textbf{concept only experiment results} in \textit{supplementary material}.}
\subsection{Dataset}
\begin{table}[]
\centering
\caption{Statics information of the Concept-7 dataset.}
\label{tab:dataset}
\begin{tabular}{ll}
\hline
\multicolumn{1}{l|}{\# of basic concepts}          & 192 \\
\multicolumn{1}{l|}{\# of basic instructions}           & 451 \\
\multicolumn{1}{l|}{\# of test concepts}           & 180 \\
\multicolumn{1}{l|}{\# of real test concepts}      & 106 \\
\multicolumn{1}{l|}{\# of fictional test concepts} & 74  \\
\multicolumn{1}{l|}{\# of test instructions}            & 515 \\ \hline
\end{tabular}
\end{table}

Most existing datasets predominantly concentrate on the classification of hallucinatory responses. To effectively evaluate our method, we introduce the {\data} dataset, which focuses on the classification of potential hallucinatory instructions. This dataset encompasses 192 basic concepts with 451 basic instructions and 180 test concepts with 515 test instructions sourced from seven expert domains. A comprehensive overview of the dataset is displayed in Table~\ref{tab:dataset}.
\subsubsection{Dataset Creation}
In the subsequent section, we delve into the specifics of dataset construction.

\noindent$\bullet$ \textbf{Concept Selection.}
Initially, we select 192 fundamental concepts, guided by the popular pages on Wikipedia\footnote{\url{https://en.wikipedia.org/wiki/Wikipedia:Popular_pages}} and domain diversity. These concepts are considered universally known for all language models and serve as a benchmark for the classification of familiar concepts.
For the test concepts, we choose 106 concepts from seven expert domains: Medical, Finance, Music, Art, Legal, Physics, and History. To maintain a balanced proportion between familiar and unfamiliar concepts, we fabricate 74 fictional concepts built upon real ones. As these fictional concepts don't exist, they're inherently deemed unfamiliar to all language models. However, as the training sources vary for each language model, it is necessary to annotate familiarity scores for the real concepts for each model under test.

\noindent$\bullet$ \textbf{Familiarity Annotation.}
To assign familiarity scores, we propose a comparison between a crafted gold explanation of concepts and each language model's generated explanation. This familiarity assessment is conducted automatically via GPT-4. We instruct GPT-4 to provide a familiarity score for the model in relation to the concept on a scale of ``1-9'', as described in \textit{supplementary material}. A threshold of ``5'' is chosen; concepts with a score exceeding ``5'' are deemed familiar, whereas those scoring less are deemed unfamiliar. Concepts scoring exactly ``5'' are manually reviewed for final labeling. Additionally, we employ Amazon Mechanical Turk\footnote{\url{https://www.mturk.com/}} to gather pure \textbf{human annotation} results for the highest-performing model as a secondary evaluation method, ensuring the effectiveness of the GPT-4 annotated results. For human-annotated results, we have three different annotators for each concept and the final label is decided based on the average score. If the average score happens to be “5”, a majority vote is applied to decide the label.

\noindent$\bullet$ \textbf{Generating Instructions}
To replicate general conversational scenarios, we use a prompt (as shown in \textit{supplementary material}) to generate three related questions for each concept via GPT-3.5. This includes two open-ended questions and one yes-or-no question. We subsequently discard questions that fail to mention the original concept to maintain strong relevance.  Instructions comprising unfamiliar concepts are regarded as hallucinatory instructions. 

\begin{table*}[t]
\vspace{0.1in}
\caption{Hallucinatory instruction classification results on the four models.}
\label{tab:major-exp}
\footnotesize
\centering
\resizebox{\textwidth}{!}{
\begin{tabular}{l|cccc|cccc|cccc|cccc}
\hline
\multirow{2}{*}{Methods} & \multicolumn{4}{c|}{Vicuna-13b-v1.3} & \multicolumn{4}{c|}{Falcon-7b-instruct} & \multicolumn{4}{c|}{mpt-7b-instruct} & \multicolumn{4}{c}{Alpaca-7b}  \\ \cline{2-17} 
                               & AUC     & ACC     & F1      & PEA    & AUC      & ACC      & F1      & PEA     & AUC     & ACC     & F1     & PEA     & AUC   & ACC   & F1    & PEA    \\ \hline
Greedy-Perplexity              & 0.867   & 0.497   & 0.651   & 0.266  & 0.697    & 0.487    & 0.553   & 0.299   & 0.639   & 0.616   & 0.000  & 0.182   & 0.701 & 0.470  & 0.614 & 0.314  \\
Greedy-AvgLogp           & 0.806   & 0.676   & 0.728   & 0.393  & 0.621    & 0.381    & 0.504   & 0.180   & 0.558   & 0.384   & 0.555  & 0.034   & 0.641 & 0.454 & 0.590 & 0.204  \\
Greedy-MinLogp           & 0.760   & 0.604   & 0.693   & 0.398  & 0.631    & 0.670    & 0.000   & 0.227   & 0.621   & 0.616   & 0.000  & 0.170   & 0.591 & 0.460 & 0.596 & 0.121  \\
Greedy-Significance            & 0.585   & 0.503   & 0.647   & 0.031  & 0.706    & 0.417    & 0.531   & 0.211   & 0.721   & 0.487   & 0.596  & 0.181   & 0.532 & 0.427 & 0.596 & -0.091 \\
Sample-BERTScore         & 0.872   & 0.779   & 0.807   & 0.640  & 0.666    & 0.416    & 0.526   & 0.180   & 0.724   & 0.497   & 0.602  & 0.299   & 0.650 & 0.517 & 0.634 & 0.196  \\
Sample-SentenceScore     & 0.831   & 0.718   & 0.757   & 0.628  & 0.656    & 0.421    & 0.527   & 0.270   & 0.736   & 0.480    & 0.590   & 0.326   & 0.653 & 0.497 & 0.621 & 0.265  \\
Forward-Inference              & 0.809   & 0.720   & 0.744   & 0.402  & 0.511    & 0.330     & 0.495   & 0.099   & 0.638   & 0.470    & 0.563  & -0.081  & 0.520  & 0.515 & 0.432 & 0.070  \\
\rowcolor{blue!25}
{\method}     & \textbf{0.927} & \textbf{0.868} & \textbf{0.854} & \textbf{0.693} & \textbf{0.926} & \textbf{0.882} & \textbf{0.822} & \textbf{0.687} & \textbf{0.921} & \textbf{0.850} & \textbf{0.817} & \textbf{0.661} & \textbf{0.918} & \textbf{0.866} & \textbf{0.848} & \textbf{0.685} \\ \hline
\end{tabular}}
\end{table*}

\subsection{Baseline Methods}
Considering that prior methodologies primarily focused on the detection of hallucinatory responses, their settings could not be directly applied in this context. Nevertheless, we strived to adjust these settings to establish the following baseline methods. Owing to space limitations, we only discuss the core concepts here, leaving comprehensive details of each baseline method for the \textit{supplementary material}. It is important to note that as our focus lies on zero-resource prevention settings, we excluded methods that necessitate external knowledge.
\begin{itemize}[leftmargin=*]
    \item Greedy-Perplexity: For each input, we utilize greedy search to generate a response and then calculate the response's perplexity. The negative perplexity score is considered the familiarity score, similar to the approach in \cite{manakul2023selfcheckgpt}.
    \item Greedy-AvgLogp: For each input, we utilize greedy search to generate a response and then calculate the response's average log token probability score, similar to the approach in \cite{manakul2023selfcheckgpt}.
    \item Greedy-MinLogp: Similar to Greedy-AvgLogp, but take the minimal probability score, akin to \cite{manakul2023selfcheckgpt}.
    \item Greedy-Significance: We first ask the model to generate an explanation response using greedy search, after which the prompt's core concepts are masked. We then regenerate the previous explanation through force decoding and compare the KL divergence of the output probability sequences between the original and masked instruction. This divergence is treated as the familiarity score.
    \item Sample-BERTScore: We sample $T_S$ responses from the prompt and evaluate the BERTScore~\cite{zhang2019bertscore} similarity between each pair of sentences. We then select the sentence with the highest average similarity score to the remaining sentences as the central sentence. The highest average similarity is treated as the familiarity score, following the method in \cite{manakul2023selfcheckgpt}.
    \item Sample-SentenceScore: Like Sample-BERTScore, we sample $T_S$ responses from the prompt and compare their sentence embedding cosine similarity, as calculated by a Sentence-BERT~\cite{reimers-2019-sentence-bert} encoder. 
    \item Forward-Inference: We directly inquire if the language model recognizes the domain-related concepts within the instruction, resembling the CoT methods~\cite{zhang2023interpretable, mundler2023self}. The likelihood of a ``Yes'' response sequence is considered the familiarity score. If the model responds with ``No'' or anything other than ``Yes'', the familiarity score is determined as one minus the probability of the response sequence.
\end{itemize}
\subsection{Tested Large Language Models}
In order to enable an in-depth comparison between different styles of instruction-aligned large language models, we have selected four distinct models for evaluation: Vicuna-13b-v1.3~\cite{zheng2023judging}, Falcon-7b-instruct~\cite{almazrouei2023falcon}, mpt-7b-instruct~\footnote{\url{https://github.com/mosaicml/llm-foundry/}}, and Alpaca-7b~\cite{taori2023alpaca}. \emph{Given our approach's requirement for constrained beam search generation control, models that exclusively offer API access were not considered in this study.}

\subsection{Evaluation Metrics}
We adopt the area under the curve (AUC), accuracy (ACC), F-score (F1), and the Pearson Correlation Coefficient (PEA) between the predicted and annotated familiarity scores.

\subsection{Results}
We begin our discussion by analyzing the hallucinatory instruction classification results presented in Table~\ref{tab:major-exp}. The table reveals two primary insights. Firstly, apart from our method, all other baseline approaches demonstrate notable performance inconsistency across the various models tested. Furthermore, the favored methods among different Language Models (LMs) significantly differ from one another.
As highlighted in the introduction, many current methods are easily influenced by model styles. As a consequence, the performance of certain methods can vary based on the models in use. This variability renders these methods less versatile across different settings.
A notable example is the Forward-Inference method. While it showcases commendable performance on Vicuna-13b-v1.3, its efficacy diminishes with other models. This observation supports the hypothesis that techniques like the CoT or prompt programming, though often capable of delivering high-quality results, are heavily reliant on the model's intrinsic CoT capacity. Since many LMs aren't specifically fine-tuned for this purpose, it restricts their widespread utility. On the other hand, Greedy-Perplexity performs well across various models but fails to detect any hallucinated instructions on mpt-7b-instruct, resulting in an F1 score of 0. This underscores the idea that even parameter-based methods are not immune to robustness issues. Similarly, other methods exhibit this same challenge.
In contrast to existing approaches, our method not only delivers superior performance but also ensures consistent results across various LMs. Additionally, the PEA correlation score demonstrates that the evaluations produced by our algorithm align closely with the familiarity scores based on gold explanations of concepts. These results underscore the robustness and reliability of our proposed approach.
\subsection{Human Evaluation Results}
\begin{table}[t]
\footnotesize
\centering
\caption{Human annotated result evaluation.}
\label{tab:human-level-exp}
\begin{tabular}{l|cccc}
\hline
\multirow{2}{*}{Methods} & \multicolumn{4}{c}{Vicuna-13b-v1.3}                               \\ \cline{2-5} 
                               & AUC            & ACC            & F1             & PEA            \\ \hline
Greedy-Perplexity              & 0.847          & 0.511          & 0.665          & 0.201          \\
Greedy-AvgLogP           & 0.781   & 0.674   & 0.730  & 0.300  \\
Greedy-MinLogp           & 0.733   & 0.493   & 0.661  & 0.264  \\
Greedy-Significance            & 0.581          & 0.517          & 0.661          & 0.082          \\
Sample-BERTScore         & 0.838   & 0.753   & 0.789  & 0.494  \\
Sample-SentenceScore     & 0.792   & 0.701   & 0.746  & 0.543  \\
Forward-Inference              & 0.766          & 0.691          & 0.762          & 0.290          \\
\rowcolor{blue!25}
{\method}     & \textbf{0.892} & \textbf{0.827} & \textbf{0.813} & \textbf{0.526} \\ \hline
\end{tabular}
\end{table}
In addition to evaluations based on GPT-4, we further utilized crowd-sourcing for the annotation of the concept familiarity scores for Vicuna-13b-v1.3 and then assessed these human-annotated results. The outcomes are delineated in Table~\ref{tab:human-level-exp}. The outcome and ranking of different methods are similar to the GPT-4 based results, proving the effectiveness of our auto-evaluation methodology utilizing GPT-4.
Finally, under the human-based evaluation, our approach still consistently exhibits superior performance across all evaluated metrics.
\subsection{Ablation Study}
\begin{table}[h]
\centering
\caption{Entity processing ablation study}
\label{tab:proce}
\begin{tabular}{l|ccc}
\hline
\multirow{2}{*}{Methods}   & \multicolumn{3}{c}{Vicuna-13b-v1.3} \\ \cline{2-4} 
                           & AUC        & ACC        & F1        \\ \hline
\rowcolor{blue!25}
{\method} & \textbf{0.927} & \textbf{0.868} & \textbf{0.854}     \\
W/O Grouping               & 0.918      & 0.856      & 0.841     \\
W/O Filtration             & 0.923      & 0.856      & 0.841     \\
W/O Ranking                & 0.926      & 0.86       & 0.845     \\ \hline
Minimal Only               & 0.902      & 0.808      & 0.767     \\
Most Infrequent Only       & 0.921      & 0.866      & \underline{0.858}     \\ \hline
\end{tabular}
\end{table}
In this section, we present an ablation study in Table~\ref{tab:proce} to examine the contribution of the proposed concept processing methods and score aggregation methods to the overall performance of our model. We first examine the different concept processing strategies. 
The following notations represent different configurations of our algorithm: (1) \textbf{W/O Grouping} denotes processing without grouping the extracted concepts.
(2) \textbf{W/O Filtering} denotes processing without filtering out common concepts.
(3) \textbf{W/O Ranking} means that concepts are not ranked based on their frequency scores. Instead, the position of the concepts within the instruction determines their order.
Results indicate that excluding any of these techniques leads to a drop in final performance. This underscores the efficacy of each proposed processing strategy.

Next, we benchmark the efficacy of other instruction-level familiarity scores aggregation techniques against our weighted averaging method: (1) \textbf{Minimal Only} selects the smallest concept familiarity score as the final outcome.
(2) \textbf{Most Infrequent Only} chooses the familiarity score of the concept with the least frequency score $f_i$ as the final result.
It is evident from the table that our proposed method demonstrates the best overall performance except for the F1 metric. This is attributable to its capability of accounting for both the importance rank of the diverse concepts and the aggregate performance. Conversely, the other two methods solely consider segments of the concept familiarity scores.

\subsection{Case Study}
\begin{figure}[h]
\centering
\includegraphics[width=\textwidth]{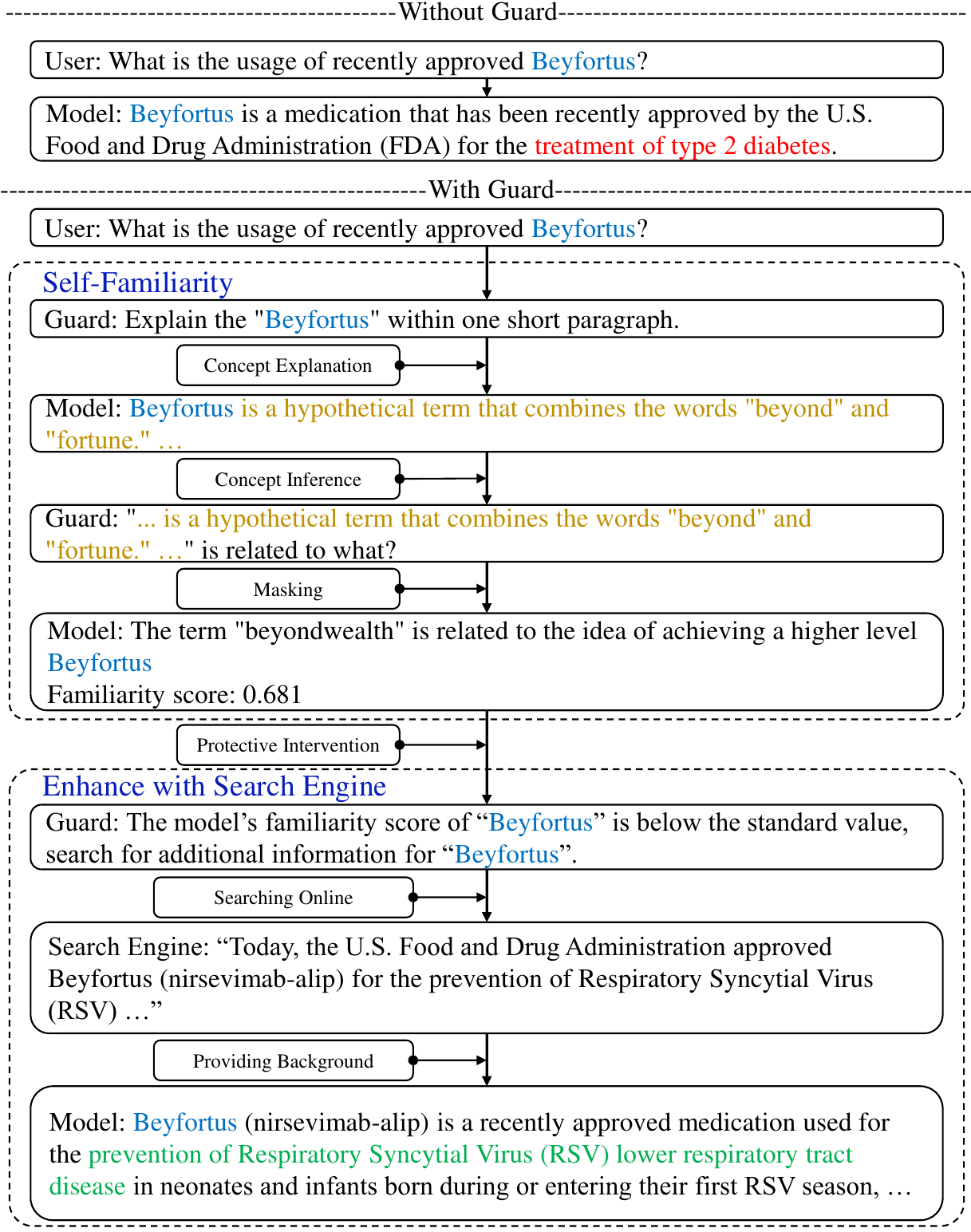}
\caption{The \textcolor{red}{Red} color denotes misinformation, while \textcolor{green}{Green} signifies correct information. \textcolor{cyan}{Blue} is the concept and \textcolor{Gold}{Gold} is the generated explanation. The tested model is Vicuna-13b-v1.3.
}
\label{fig:case}
\end{figure}
In this section, we perform a real-world case study on the medical domain to demonstrate the applicability of our algorithm in preventing hallucinations.
The details can be found in Figure~\ref{fig:case}. 
\textit{``Beyfortus''}\footnote{\url{https://www.fda.gov/news-events/press-announcements/fda-approves-new-drug-prevent-rsv-babies-and-toddlers}} is newly approved drug.
We initially examine the response in the absence of our algorithm in ``Without Guard'', wherein the model unhesitatingly disseminates misinformation. This kind of misinformation is challenging to detect unless one proactively seeks the underlying background information.
Subsequently, in ``With Guard'', our algorithm serves as a guard, assisting us in evaluating the model's comprehension of \textit{``Beyfortus''} utilizing the {\method}. It is evident that the model encounters considerable difficulty in generating a response associated with \textit{``Beyfortus''} based on the masked explanation. This is because the model lacks an intrinsic learned connection between the concept and the fabricated explanation.
Furthermore, we can readily address these issues by introducing background knowledge of unfamiliar concepts. In the subsequent step, the search engine is activated to retrieve information related to \textit{``Beyfortus''} as background data, and the model is then capable of rendering the correct response. These results suggest that our approach is not only potent in prevention but can also offer great interpretability and serve as a valuable tool in correcting hallucinated responses.

\section{Conclusion}
We have introduced a cutting-edge pre-detection mechanism for potential hallucination instruction, which we refer to as {\method}. Our approach leverages {\inference} to assess the model's quality of concept explanation, thereby determining the model's level of understanding. {\method} consistently achieves state-of-the-art results in the pre-detection of hallucinatory instruction across four distinct language models using only self-evaluation under the zero-resource setting.
Additionally, our method demonstrates superior interpretability by identifying the particular concept that led to the hallucination. This unique feature enables the integration of our method with post-detection and correction techniques, enhancing its versatility.
In future work, we plan to investigate how to evaluate the understanding of more granular sub-concepts to further refine the precision of the current algorithm.

\bibliography{iclr2021_conference}
\bibliographystyle{iclr2021_conference}

\appendix
\section{Prompts}
\begin{figure}[htpb]
\centering
\fbox{\includegraphics[width=\linewidth]{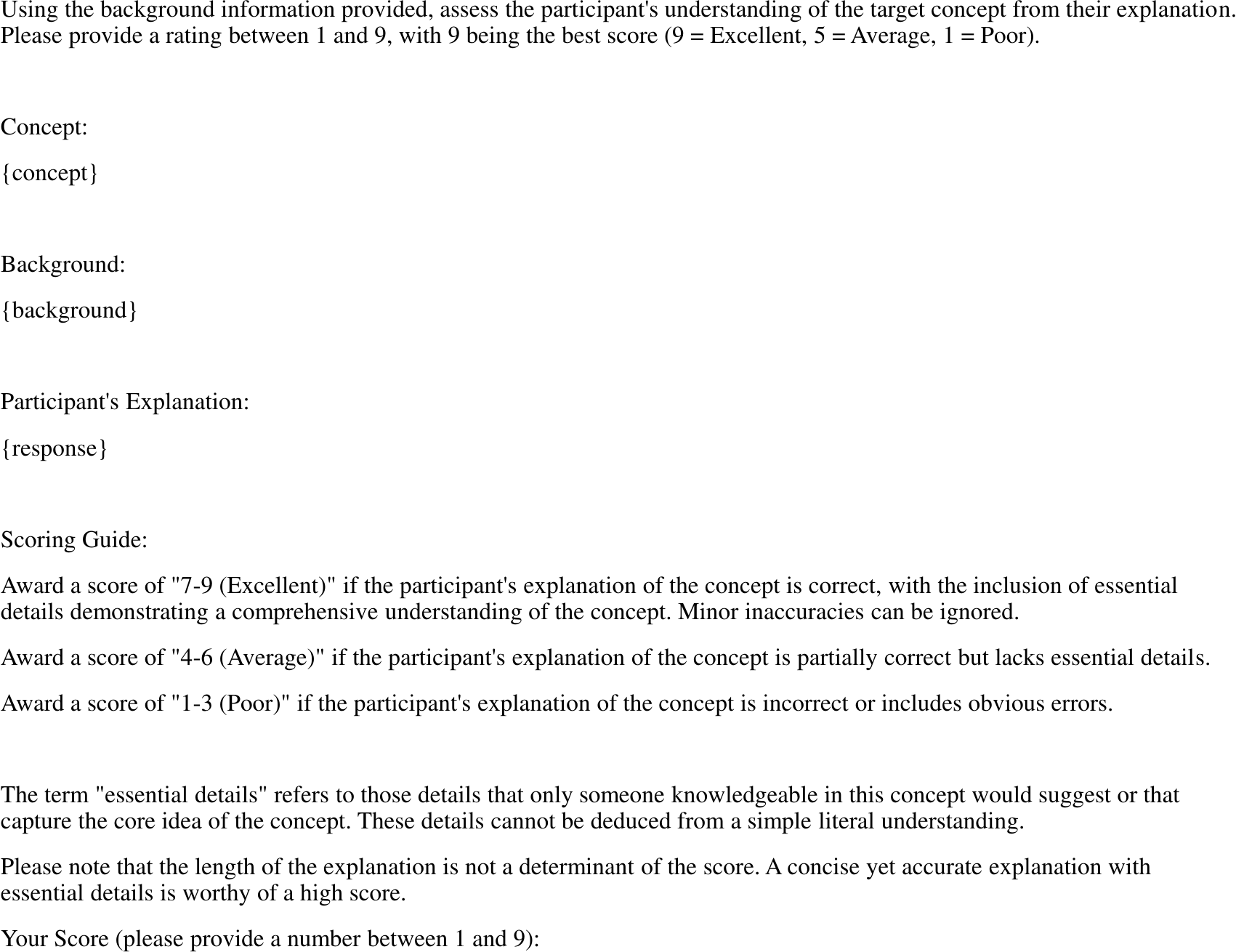}}
\caption{Scoring prompt used by GPT-4 and human annotators for annotating the familiarity score.}
\label{fig:example}
\end{figure}
\begin{figure}[htpb]
\centering
\fbox{\includegraphics[width=\linewidth]{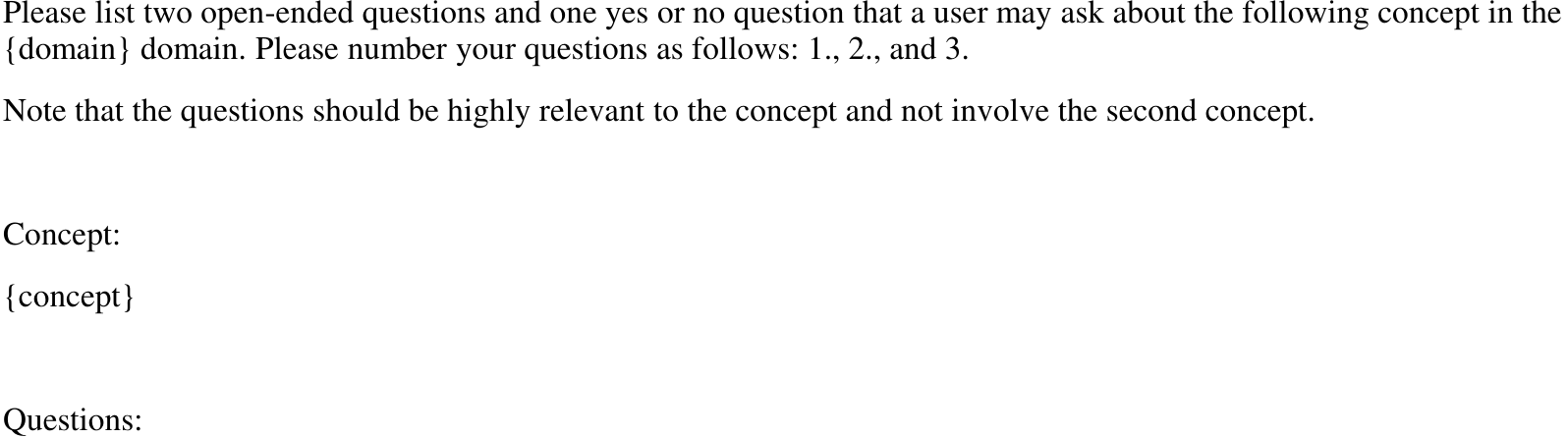}}
\caption{Question generation prompt used by GPT-3.5 for generating the instructions.}
\label{fig:example}
\end{figure}
\begin{figure}[htpb]
\centering
\fbox{\includegraphics[width=\linewidth]{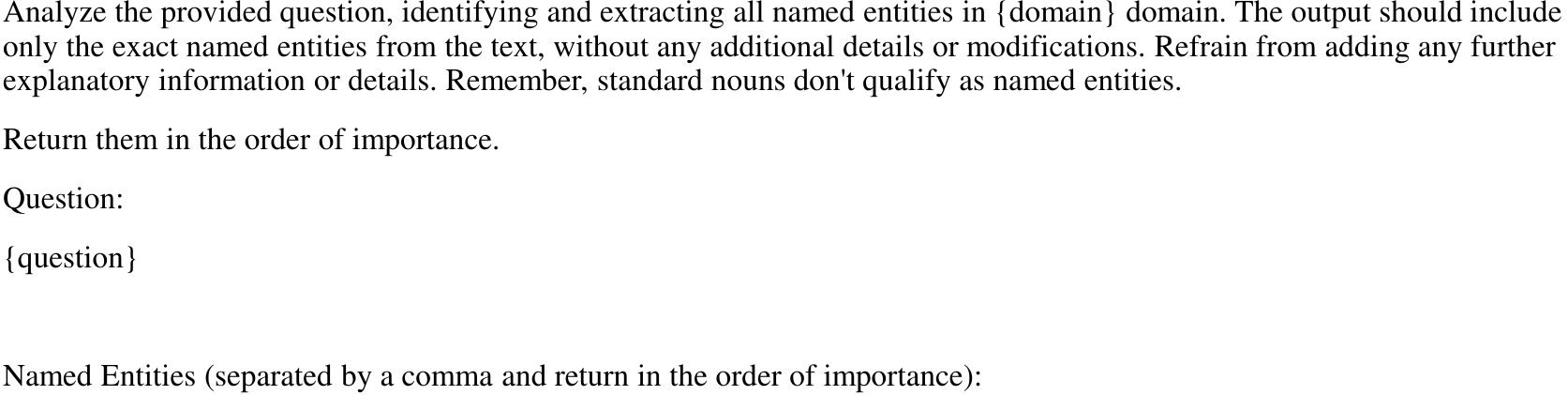}}
\caption{Named entity extraction prompt used by GPT-3.5 for extracting entities from the general instructions.}
\label{fig:example}
\end{figure}
\section{Implementation Details}
For all baselines, we limit the maximum length of the response to 200 tokens. For sampling methods, the number of samples, $T_s$, is set to 10. For our method, we set $l_F$ to 200 and $l_B$ to 15, and the search beam size $T_B$ is set to 30. We use ``...'' as the mask token since not every language model has the ''[MASK]'' token and we select the top 10,000 words in Wiktionary as the ``common words''. The decay ratio $r$ is set to 2 and the normalization factor $H$ is set to 100. Finally, We use GPT-3.5 to extract critical concepts from the sampled questions as a zero-resource domain named entity extractor. The prompt used for named entity extraction is displayed in \textit{supplementary material}.
The following prompts are employed to perform \textbf{Concept Explanation} and conduct \textbf{Concept Inference} for concept only and general-level experiments:
\begin{itemize}[leftmargin=*]
    \footnotesize
    \item \textbf{Concept Explanation} (concept only) $P_{E}$: \textit{Explain the \{concept\} in the \{domain\} domain within one short paragraph.}
    \item \textbf{Concept Explanation} $P_{E}$: \textit{Explain the ``\{concept\}'' within one short paragraph.}
    \item \textbf{Concept Inference} $P_{I}$: \textit{``\{masked explanation\}'' is related to what?}
\end{itemize}
For all settings and methods, we employ 192 basic concepts-related instructions to estimate the hallucination threshold $h$ for each method and model. We use bootstrap sampling~\cite{efron1994introduction} to sample the 95\% threshold interval of the scores of basic concepts and use the midpoint of the interval as the threshold.
Finally, all methods were tested on a PyTorch-based system running Ubuntu 20.04, equipped with 128 GB of memory and two NVIDIA A100 GPUs.
To increase the reproducibility, we fix the random seed to 42 to run the experiment and use the max probability to prompt the LM to generate responses unless the sampling is required to avoid randomness.
\section{Baselines}
\subsection{Greedy-Perplexity}
A model with higher perplexity typically indicates greater uncertainty in its responses, which can suggest potential hallucinations. Initially, the response is generated using Greedy Search:
\begin{equation}
R = \mbox{GreedySearch}(\mbox{LM}(P_T))
\end{equation}
Subsequently, the perplexity score of response $R$ is determined:
\begin{equation}
s = -\mbox{Perplexity}(R)
\end{equation}
The inverse of this perplexity score is considered the faithfulness score.
\subsection{Greedy-AvgLogp}

Lower probability values generally signify that the model is more uncertain, hinting at a potential hallucination. The response is generated using Greedy Search:
\begin{equation}
    R = \mbox{GreedySearch}(\mbox{LM}(P_T))
\end{equation}
The probability sequence of \(R\) is then determined:
\begin{equation}
    \mathbf{P} = \mbox{ProSeq}(R)
\end{equation}
The mean log value of \(\mathbf{P}\) is computed as:
\begin{equation}
    s = \mbox{Avg}(\mbox{log}(\mathbf{P}))
\end{equation}
This average, $s$, is considered the faithfulness score.
\subsection{Greedy-MinLogp}
This method mirrors the Greedy-AvgLogp but employs the minimal score. The response generation and probability sequence determination are consistent:
\begin{equation}
    R = \mbox{GreedySearch}(\mbox{LM}(P_T))
\end{equation}
\begin{equation}
    \mathbf{P} = \mbox{ProSeq}(R)
\end{equation}
The minimum log value of \(\mathbf{P}\) is:
\begin{equation}
    s = \mbox{Min}(\mbox{log}(\mathbf{P}))
\end{equation}
Here, $s$ denotes the faithfulness score.
\subsection{Greedy-Significance}
A small difference between the generated explanation and an explanation generated without conditions can hint that the concept entities have minimal impact on the final outcome, signaling potential hallucinations. We Use the same concept entity extraction technique as our method to extract the concept entities. Then, the response is generated using Greedy Search:
\begin{equation}
    R = \mbox{GreedySearch}(\mbox{LM}(P_F\oplus e_i))
\end{equation}
Simultaneously, the probability sequence of $R$ is determined:
\begin{equation}
    \mathbf{P} = \mbox{ProSeq}(R)
\end{equation}
To regenerate the prior output $R$ under an unconditional setting (where concepts are masked), the force decode is applied, obtaining:
\begin{equation}
    \mathbf{P}^* = \mbox{ForceDecode}(\mbox{LM}(P_F\oplus\mbox{[MASK]}), R)
\end{equation}
We then compute the difference between conditional probability sequence $\mathbf{P}^*$ and unconditional probability sequence $\mathbf{P}$:
\begin{equation}
    s = \mbox{KL-divergence}(R^*, R)
\end{equation}
\subsection{Sample-BERTScore}
If there is high similarity between generated responses, it indicates the sampled responses are consistent. Conversely, diverse responses suggest the model's uncertainty about the instruction. 
We first sample $T_s$ responses based on the instruction:
\begin{equation}
    [R_1,\cdots, R_{T_s}] = \mbox{Sample}(\mbox{LM}(P_T))
\end{equation}
Next, we compute the similarity between any two responses:
\begin{equation}
    s^{sim}_{i,j} = \mbox{BERTScore}(R_i, R_j)
\end{equation}
For each response, we calculate the aver1age similarity to all the responses:
\begin{equation}
    s^{sim}_{i} = \frac{1}{m}\sum_{j=1}^m s^{sim}_{i,j}
    \label{eq:1}
\end{equation}
Each response's average similarity to all others is determined, with the highest average similarity representing the final score:
\begin{equation}
    s = \mbox{max}(s^{sim}_{1},\cdots,s^{sim}_{T_s})
    \label{eq:2}
\end{equation}
\subsection{Sample-SentenceScore}
Similar to the Sample-BERTScore, the only difference is that we use the sentence transformer to obtain the sentence embedding of each sampled response $R_i$:
\begin{equation}
    \mathbf{r}_i = \mbox{SentenceTransformer}(R_i)
\end{equation}
Next, we utilize the Cosine Similarity to obtain the similarity score between any two responses based on the sentence embedding:
\begin{equation}
    s^{sim}_{i,j} = \mbox{CosineSimilarity}(\mathbf{r}_i, \mathbf{r}_j)
\end{equation}
The following parts are identical to Eq.\eqref{eq:1}-\eqref{eq:2}.
\subsection{Direct-Inference}
We directly ask if the model is familiar with the concepts through the following prompts:
\begin{itemize}
    \item $P_D$ (concept only): \textit{"Are you familiar with the \{concept\} in \{domain\}? Answer yes or no."}
    \item $P_D$: \textit{"Are you familiar with all the \{domain\} concepts in "\{instruction\}"? Answer yes or no."}
\end{itemize}
\begin{equation}
    R = \mbox{GreedySearch}(\mbox{LM}(P_D))
\end{equation}
Based on the response, we calculate the hallucination score based on the overall probability score of the response $R$:
\begin{equation}
s = 
\begin{cases}
\mbox{Prob}(R),\ \mbox{"Yes" in } R\\
1-\mbox{Prob}(R),\ \mbox{else}
\end{cases}
\end{equation}
Note that keywords are considered in their lowercase, uppercase, and capitalized forms.
\begin{table*}[h]
\caption{Results of concept only on the four models.}
\label{tab:concept only-exp}
\footnotesize
\centering
\resizebox{\textwidth}{!}{\begin{tabular}{l|cccc|cccc|cccc|cccc}
\hline
\multicolumn{1}{c|}{\multirow{2}{*}{Methods}} & \multicolumn{4}{c|}{Vicuna-13b-v1.3} & \multicolumn{4}{c|}{Falcon-7b-instruct} & \multicolumn{4}{c|}{mpt-7b-instruct} & \multicolumn{4}{c}{Alpaca-7b}   \\ \cline{2-17} 
\multicolumn{1}{c|}{}                         & AUC     & ACC    & F1     & Pearson  & AUC     & ACC     & F1      & Pearson   & AUC     & ACC    & F1     & Pearson  & AUC   & ACC   & F1    & Pearson \\ \hline
Greedy-Perplexity                             & 0.651   & 0.511  & 0.645  & 0.216    & 0.718   & 0.467   & 0.551   & 0.409     & 0.536   & 0.617  & 0.000  & -0.053   & 0.653 & 0.494 & 0.609 & 0.162   \\
Greedy-AvgLogp                             & 0.784   & 0.478  & 0.647  & 0.467    & 0.747   & 0.550   & 0.571   & 0.439     & 0.652   & 0.383  & 0.554  & 0.234    & 0.673 & 0.433 & 0.605 & 0.280   \\
Greedy-MinLogp                             & 0.724   & 0.478  & 0.647  & 0.394    & 0.725   & 0.422   & 0.519   & 0.427     & 0.758   & 0.383  & 0.554  & 0.419    & 0.659 & 0.433 & 0.605 & 0.270   \\
Greedy-Significance                           & 0.593   & 0.556  & 0.675  & 0.042    & 0.687   & 0.583   & 0.611   & 0.033     & 0.597   & 0.528  & 0.585  & -0.039   & 0.454 & 0.45  & 0.599 & -0.260  \\
Sample-BERTScore                              & 0.920   & 0.678  & 0.748  & 0.718    & 0.863   & 0.600   & 0.617   & 0.711     & 0.802   & 0.528  & 0.615  & 0.393    & 0.876 & 0.722 & 0.742 & 0.700   \\
Sample-SentenceScore                          & 0.883   & 0.639  & 0.726  & 0.788    & 0.850   & 0.556   & 0.592   & 0.634     & 0.833   & 0.606  & 0.657  & 0.538    & 0.809 & 0.639 & 0.700 & 0.666   \\
Forward-Inference                             & 0.902   & 0.722  & 0.769  & 0.255    & 0.564   & 0.356   & 0.508   & 0.005     & 0.568   & 0.556  & 0.474  & 0.206    & 0.783 & 0.722 & 0.662 & 0.486   \\ \rowcolor{blue!25}
{\method}                            & \textbf{0.966} & \textbf{0.928} & \textbf{0.921} & \textbf{0.844} & \textbf{0.968} & \textbf{0.911} & \textbf{0.864} & \textbf{0.772} & \textbf{0.935} & \textbf{0.911} & \textbf{0.892} & \textbf{0.631} & \textbf{0.905} & \textbf{0.883} & \textbf{0.863} & \textbf{0.680} \\ \hline
\end{tabular}}
\end{table*}
\section{Concept Only Evaluation}
This scenario aims to evaluate the {\inference} stage solely. The results can be found in Table~\ref{tab:concept only-exp} and the Table~\ref{tab:human-level-exp}. The target concept is directly provided and the concept explanation prompt $P_F$ \textit{``Explain the \{concept\} in the \{domain\} domain within one short paragraph.''} is uniformly applied to the baselines as the input instruction, which is also the concept explanation prompt of our {\inference}. In this scenario, our method consistently achieves state-of-the-art (SOTA) performance. This demonstrates that even in less-noisy situations, our approach still surpasses the baseline methods.
\begin{table}[h]
\footnotesize
\centering
\caption{Human annotated result evaluation on concept only.}
\label{tab:human-level-exp}
\begin{tabular}{l|cccc}
\hline
\multicolumn{1}{c|}{\multirow{2}{*}{Methods}} & \multicolumn{4}{c}{Vicuna-13b-v1.3}                               \\ \cline{2-5} 
\multicolumn{1}{c|}{}                         & AUC            & ACC            & F1             & Pearson        \\ \hline
Greedy-Perplexity                             & 0.681          & 0.528          & 0.661          & 0.189          \\
Greedy-AvgLogp                                & 0.765          & 0.494          & 0.662          & 0.361          \\
Greedy-MinLogp                                & 0.712          & 0.494          & 0.662          & 0.240          \\
Greedy-Significance                           & 0.600          & 0.572          & 0.691          & 0.071          \\
Sample-BERTScore                              & 0.884          & 0.683          & 0.755          & 0.496          \\
Sample-SentenceScore                          & 0.841          & 0.644          & 0.733          & 0.613          \\
Forward-Inference                             & 0.895          & 0.717          & 0.767          & 0.362          \\\rowcolor{blue!25}
{\method}                    & \textbf{0.920} & \textbf{0.889} & \textbf{0.881} & \textbf{0.678} \\ \hline
\end{tabular}
\end{table}

\end{document}